\documentclass[letterpaper, 10 pt, conference]{ieeeconf}  

\IEEEoverridecommandlockouts                              

\usepackage{times}

\usepackage[ruled,vlined]{algorithm2e}
\usepackage[noend]{algcompatible}

\usepackage{amsmath}
\usepackage{amsfonts}
\usepackage{breqn}
\usepackage{multirow,hhline,graphicx,array}
\graphicspath{ {./images/} }
\usepackage{multicol}
\usepackage[bookmarks=true]{hyperref}
\usepackage{subfiles}
\usepackage{caption}
\usepackage{footnote}
\usepackage{gensymb}
\usepackage{float}
\usepackage[flushleft]{threeparttable}

\usepackage{xcolor}

\title{\LARGE \bf
SemCal: Semantic LiDAR-Camera Calibration using Neural Mutual Information Estimator
}

\author{Peng Jiang$^{1}$, Philip Osteen$^{2}$, and Srikanth Saripalli$^{1}$
\thanks{$^{1}$J. Mike Walker '66 Department of Mechanical Engineering, 
Texas A\&M University, College Station, TX 77843, USA
        {\tt\small \{maskjp, ssaripalli\}@tamu.edu}}%
\thanks{$^{2}$DEVCOM Army Research Laboratory (ARL), Adelphi, MD 20783, USA
        {\tt\small philip.r.osteen.civ@mail.mil}}%
}

\begin{document}

\maketitle
\thispagestyle{empty}
\pagestyle{empty}

\begin{abstract}
This paper proposes SemCal: an automatic, targetless, extrinsic calibration algorithm for a LiDAR and camera system using semantic information. We leverage a neural information estimator to estimate the mutual information (MI) of semantic information extracted from each sensor measurement, facilitating semantic-level data association. By using a matrix exponential formulation of the $se(3)$ transformation and a kernel-based sampling method to sample from camera measurement based on LiDAR projected points, we can formulate the LiDAR-Camera calibration problem as a novel differentiable objective function that supports gradient-based optimization methods. We also introduce a semantic-based initial calibration method using 2D MI-based image registration and Perspective-n-Point (PnP) solver. To evaluate performance, we demonstrate the robustness of our method and quantitatively analyze the accuracy using a synthetic dataset. We also evaluate our algorithm qualitatively on an urban dataset (KITTI360) and an off-road dataset (RELLIS-3D) benchmark datasets using both hand-annotated ground truth labels as well as labels predicted by the state-of-the-art deep learning models, showing improvement over recent comparable calibration approaches. 
\end{abstract}

\section{Introduction}
Camera and Light Detection and Ranging (LiDAR) sensors are essential components of autonomous vehicles which have complementary properties. A camera can provide high-resolution color information but is sensitive to illumination and lacks direct spatial measurement. A LiDAR can provide accurate spatial information at long ranges and is robust to illumination changes, but its resolution is much lower than the camera and does not measure color. Fusing the measurements of these two sensors allows autonomous vehicles to have an improved spatial and visual understanding of the environment. In order to combine the data from different sensor modalities, it is essential to have an accurate transformation between the coordinate systems of the sensors. Therefore, calibration is a crucial first step for multi-modal sensor fusion.

In recent years, many LiDAR-camera calibration methods have been proposed. These methods can be categorized based on whether they are online \cite{Nagy2019}\cite{Wang2020}\cite{Zhu2020}\cite{Lv2020}\cite{Yuan2020}\cite{Chien2016} or offline \cite{Mishra2020},\cite{Mishra2020a} \cite{Owens2015} as well as whether they require a calibration target \cite{Mishra2020},\cite{Mishra2020a} or not \cite{Lv2020}\cite{Yuan2020}\cite{Chien2016}\cite{Zhao2021}. Target-based methods require carefully designed targets \cite{Zhang2004}, \cite{Pusztai2017}, \cite{Owens2015} or otherwise controlled environments while targetless methods use data from the natural environment to perform calibration. Target-based offline methods can provide accurate results and are easier to evaluate due to experimental controls, however unpredictable environmental conditions (e.g., heat, vibration, impact, etc.) can degrade calibration accuracy over time, and therefore sensors should also be calibrated online during operation \cite{Levinson2013}.

In this paper, we develop an effective targetless method for such an online multi-sensor calibration system. Targetless calibration algorithms can be categorized into whether they are non-learning-based methods or learning-based methods. Learning-based methods train neural network models to directly predict the calibration parameters \cite{Lv2020},\cite{Schneider2017},\cite{Zhao2021},\cite{Yuan2020}. Meanwhile, most traditional methods are non-learning-based methods that manually define features of each sensor measurement, assign association between the features, and  minimize or maximize a metric that measures the relationship between the two sets of features \cite{Mishra2020},\cite{Mishra2020a},\cite{Kang2020},\cite{Taylor2015},\cite{Pusztai2017},]\cite{Zhao2021}. Among the most common features employed are edges\cite{Mishra2020},\cite{Mishra2020a},\cite{Kang2020},\cite{Iyer2018}, gradients\cite{Taylor2015}, and semantics\cite{Nagy2019,Zhu2020,Wang2020}. Among these, semantic information is a higher-level feature that is available from human annotations or learning-based semantic segmentation model predictions \cite{Zhang2019}, \cite{Yu2018}. Semantic understanding is essential for autonomous systems, and several methods have been proposed to utilize semantic information to calibrate LiDAR and camera \cite{Nagy2019}\cite{Zhu2020}\cite{Wang2020},\cite{Ma2021}.

In this paper, we seek to achieve online target-free calibration from LiDAR and camera measurements, presenting a novel application of mutual information to perform optimization. Mutual information is widely used in a variety of fields such as medical image registration \cite{Oliveira2014, Nan2020}. Pandey et al.\cite{Pandey2015} first proposed applying mutual information to LiDAR-camera calibration. Their approach considers the sensor-measured surface intensities (reflectivity for LiDAR and grayscale intensity for camera) as two random variables and maximizes the mutual information between them. Inspired by \cite{Nan2020}, we use mutual information neural estimate (MINE) \cite{Belghazi2018} to estimate the mutual information. Instead of estimating the mutual information directly from the raw sensor data, we model the mutual information between the dense semantic labels derived from the data. The fundamental challenge of multi-sensor calibration is performing cross-modal data association; if we can estimate comparable semantic labels for each modality, then we should be able to perform data association using these semantics. We also use matrix exponential to compute transformation matrix and use the kernel-based sampler to sample points from images based on projected LiDAR. All together, this approach allow us to propose a fully differentiable LiDAR camera calibration framework based on semantic information and implement the algorithm using popular Deep learning libraries. Note that while we leverage deep neural networks to perform semantic segmentation, we do not directly perform regression on the calibration parameters themselves.
The major contributions of this paper can be summarized as follows:
\begin{itemize}
  \item We propose an algorithm for automatic, targetless, extrinsic calibration of a LiDAR and camera system using semantic information.
  \item We show how to make the objective function fully differentiable and optimize it using the gradient-descent method. 
  \item We introduce a semantic-based initial calibration method using 2D MI-based image registration. 
  \item We evaluate our method on a synthetic dataset from Carla simulator \cite{Dosovitskiy17}, 
 \item We experiment our method on the real-world datasets: KITTI360 dataset \cite{Xie2016CVPR}, and RELLIS-3D dataset \cite{jiang2020rellis3d}  using both the ground truth labels and label predictions from the state-of-the-art deep learning models. 
\end{itemize}
\section{Related Works}
Traditional 3D-LiDAR and camera calibration methods can be categorized into off-line, non-learning and target-based methods. This type of method is inspired from target based 2D-LiDAR camera extrinsic calibration methods \cite{Wasielewski1995}\cite{Gomez-Ojeda2015} and requires known objects in the sensor's common Field of View (FoV). The features are typically geometric features of the objects such as the edge \cite{Kang2020},\cite{Iyer2018}, plane\cite{Mishra2020},\cite{Mishra2020a}\cite{Pusztai2017}.

In this paper, we focus on targetless 3D-LiDAR camera calibration methods. Targetless methods can be sorted into non-learning and learning-based methods. Compared with feature-metric based methods, learning-based methods don't require manually defined features but rely on neural network models to learn useful features by themselves \cite{Lv2020},\cite{Schneider2017},\cite{Zhao2021},\cite{Yuan2020}. However, learning-based methods need a dataset with ground truth calibration parameters in order to train the model, which is not always available in some scenarios. Furthermore, neural network models have difficulty transferring from one dataset to another \cite{Iyer2018,Tan2018}.  

Targetless non-learning based methods also employ pre-defined features such as edges\cite{Levinson2013},\cite{Kang2020}, gradients \cite{Taylor2015} surface reflectivity and semantic information. Nagy et al. \cite{Nagy2019} use a structure from motion (SfM) pipeline to generate points from image and register with LiDAR to create a basis calibration and refine based on semantic information. Zhu et al. \cite{Zhu2020} use semantic masks of image and construct height map to encourage laser points to fall on the pixels labeled as obstacles. The methods mentioned above only use semantic information from images and also define other features to help calibration. Wang et al. \cite{Wang2020} proposed a new metric to calibrate LiDAR and camera by reducing the distance between the misaligned points and pixels based on semantic information from both image and point cloud. 
\section{Methology}
\subsection{Algorithm}

The following formula describes the transformation relationship of a point $p^L_i$ from the LiDAR local coordinate system to the camera projection plane:
\begin{equation}
\begin{bmatrix}p^{uv}_{i}\\1\end{bmatrix}=\mathrm{K}\begin{bmatrix}
1 & 0 & 0 & 0 \\
0 & 1 & 0 & 0 \\
0 & 0 & 1 & 0
\end{bmatrix}\begin{bmatrix}
\mathrm{R} & \mathrm{t} \\
0 & 1
\end{bmatrix} \begin{bmatrix}p^L_i\\1\end{bmatrix}
\label{eq:lidarcam}
\end{equation}

\begin{itemize}
    \item $p_i^L=\begin{bmatrix}x_{i}^L&y_{i}^L&z_{i}^L\end{bmatrix}^T$ represents the coordinate of point $p_i$ in Lidar local frame;
    \item $t$ represents the $3\times1$ translation vector;
    \item $R$ represents the $3\times3$ rotation matrix;
    \item $K$ represents the $3\times3$ intrinsic matrix of camera;  
    \item $p_i^{uv}=\begin{bmatrix}u_{i}&v_{i}\end{bmatrix}^T$ represents the coordinate of point $p_i$ in Camera projection plane;
\end{itemize}
This paper focuses on the extrinsic calibration between LiDAR and camera; therefore, we assume that the camera and LiDAR have been intrinsically calibrated. The extrinsic calibration parameters are given by $R$ and $t$ in Eq.\ref{eq:LiDARcam}. In addition to point cloud and image coordinates, we also assume that dense semantic labels of point cloud $L^L$ and image $L^C$ data are available, either from human annotation or a learned model. 

We consider the semantic label value of each point cloud and its corresponding image pixel as two random variables $X$ and $Y$. The mutual information of the two variables should have the maximum value when we have the correct calibration parameters. In order to perform calibration, we  need to perform the following three operations:
\begin{enumerate}
    \item \textbf{Transformation Matrix Computation}: $P^{uv} = \text{Proj}(P^{L},R,t)$ projects 3D points from LiDAR coordinate to camera coordinate;
    \item \textbf{Image Samping}: $\widetilde{L}^{C} = \text{Sample}(L^{C},P^{uv})$ samples semantic label values from image labels based on projected LiDAR coordinates.
    \item \textbf{Mutual Information Estimation}: $I(X,Y) = \text{MI}(\widetilde{L}^{C},\widetilde{L}^{L})$ estimates mutual information based on the samples from the semantic labels of LiDAR points and corresponding image pixels.
\end{enumerate}
 Therefore, our full optimization objective function can be written as Eq.\ref{eq:opt1}
\begin{equation}
    R,t=\arg \max _{R,t}\text{MI}(\text{Sample}(L^{C},\text{Proj}(P^{L},R,t)))
\label{eq:opt1}
\end{equation}
\subsection{Optimization}
The cost function Eq.\ref{eq:opt1} is maximized at the correct value of the rigid-body transformation parameters between LiDAR and camera. Therefore, any optimization technique that iteratively converges to the global optimum can be used, although we prefer gradient based optimization methods for their fast convergence properties. Here, we present how to make the cost function fully differentiable, which allows us to optimize Eq.\ref{eq:opt1} with a gradient-based optimization method. In the remaining part of this section, we describe how to make transformation matrix computation, image sampling and mutual information estimation differentiable, with the full algorithm given in Algorithm.\ref{alg:3d}.
\subsubsection{Transformation Matrix Computation}
$P^{uv} = \text{Proj}(P^{L},R,t)$ involves rigid 3D transformation $T$ which can be represented as the matrix exponential ($ T = \exp{(H)}$). And the matrix $H$ can be parameterized as a sum of weighted basis matrix of the Lie algebra $se(3)$ ($ H = \sum_{i=0}^{5} {v_iB_i}$) as described in \cite{Wachinger2013}. Therefore, we can represent $T$ with 6 parameters (see Eq.\ref{eq:expT}) and apply standard partial derivative to compute the gradient while optimizing. 
\begin{equation}
T  = \begin{bmatrix}R&t\\0&1\end{bmatrix} =\exp{(\sum_{i=0}^{5} {v_iB_i})}= \sum_{j=0}^{\infty } { \frac{(\sum_{i=0}^{5} {v_iB_i})^n }{n!} }
\label{eq:expT}
\end{equation}

\subsubsection{Image Sampling}
Differentiable image sampling is widely used in deep learning for computer vision \cite{Jaderberg2015}. The sampling operation can be written as
\begin{equation}
    \widetilde{l}_{i}^{C}=\sum_{h}^{H} \sum_{w}^{W} l_{h w}^{C} k\left(u_{i}-h; \Phi_{x}\right) k\left(v_{i}-w ; \Phi_{y}\right) 
\end{equation}
where $\Phi_{x}$ and $\Phi_{y}$ are the parameters of a generic sampling kernel $k()$ which defines the image interpolation (e.g. bilinear), $l_{h w}^{C}$ is the label value at location $(h,w)$ of the image semantic label. $\widetilde{l}_{i}^{C}$ is the corresponding semantic label of point $p_i$ after projecting on image plane.
\begin{algorithm}
\SetAlgoLined
\SetKwData{Left}{left}\SetKwData{This}{this}\SetKwData{Up}{up}
\SetKwData{Repeat}{repeat}\SetKwData{Until}{Until}
\SetKwFunction{Union}{Union}\SetKwFunction{FindCompress}{FindCompress}
\SetKwInOut{Input}{input}\SetKwInOut{Output}{output}
\Input{Lidar Point Cloud $P$, Point Cloud Labels $L^{P}$, Image Labels $L^{C}$, Camera Intrinsic Matrix $K$, Initial Transformation Matrix $T_{init}$}
\Output{Transformation Matrix $T$}
 Use $T_{init}$ to initialize $v_i$\; 
 Use random initialization for MINEnet parameters $\theta$\;
 Initialize learning rate $\alpha$, $\beta$, optimizer and learning rates scheduler\;
 \While{not converge}{
    $T = \sum_{i=0}^{5} {v_iB_i}$\;
    Sample $b$ minibatch points $P_{b}$ and labels $L_{b}^{P}$ from $P$ and $L^{P}$\;
    $P_{b}^{u,v}, = \text{Proj}(P_{b},T,K)$\;
    $\widetilde{L}_{b}^{C} = \text{Sample}(L^{C},P_{b}^{u,v})$\;
    $MI = \text{MINE}(L_{b}^{P},\widetilde{L}_{b}^{C}$)\;
    Update MIME parameter: $\theta += \alpha\bigtriangledown_{\theta}\text{MI}$\;
    Update matrix exponential parameters: $v += \beta\bigtriangledown_{v}\text{MI}$\;
 }
 Return $T = \sum_{i=0}^{5} {v_iB_i}$\;
\caption{3D Calibration}
\label{alg:3d}
\end{algorithm}
\subsubsection{Mutual Information Estimation}
Mutual information is a fundamental quantity for measuring the relationship between random variables. Traditional approaches are non-parametric (e.g., binning, likelihood-ratio estimators based on support vector machines, non-parametric kernel-density estimators), which are not differentiable. Several mutual information neural estimators have been proposed in recent works \cite{Belghazi2018, Mukherjee2019, KumarMondal}. In our implementation,  we use MINE\cite{Belghazi2018} to estimate mutual information. This method uses the Donsker-Varadhan (DV) duality to represent MI as
\begin{equation}
 \widehat{I(X ; Y)}_{n}=\sup _{\theta \in \Theta} \mathbb{E}_{P_{(X,Y)}}\left[F_{\theta}\right]-\log \left(\mathbb{E}_{P(X) P(Y)}\left[e^{F_{\theta}}\right]\right)
\label{eq:mime}
\end{equation},
where $P(X,Y)$ is the joint density for random variables $X$ and $Y$ and $P(X)$ and $P(Z)$ are marginal densities for $X$ and $Y$. $F_{\theta}$ is a function parameterized by a neural network, where $\theta$ are the parameters of the neural network.
\begin{figure*}
  \centering
  \includegraphics[width=\textwidth]{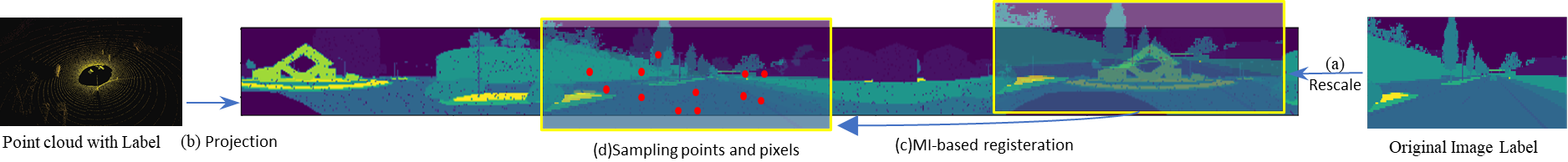}
  \caption{Initial Calibration Procedure: (a) Zoom the camera image label; (b) Project LiDAR label into 2D cylinder plane; (c) Register 2D semantic Images; (d) Sample points and pixels from the overlapped area of the two semantic labels.}
  \label{fig:initial}
\end{figure*}
\begin{algorithm}
\SetAlgoLined
\SetKwData{Left}{left}\SetKwData{This}{this}\SetKwData{Up}{up}
\SetKwData{Repeat}{repeat}\SetKwData{Until}{Until}
\SetKwFunction{Union}{Union}\SetKwFunction{FindCompress}{FindCompress}
\SetKwInOut{Input}{input}\SetKwInOut{Output}{output}
\Input{Lidar Point Cloud $P$, Point Cloud Labels $L^{P}$, Lidar Field of View $FoV_{H}^{L},FoV_{V}^{L}$, Lidar Channel Number $H^{L}$ and Ring Point Number $W^{L}$  Image Labels $L^{I}$,  Camera Field of View $FoV_{H}^{C}, FoV_{V}^{C}$, Image Height $H^{I}$ and Width $W^{I}$ }
\Output{Initial Transformation Matrix $T_{init}$}
$L^{P}_{cy} = \text{SphericalProj}(P,L^{P},H^{L},W^{L})$\;
$W_{z}^{I},H_{z}^{I} =  \frac{W^{L}}{Fov_{V}^{L}}FoV_{V}^{C}, \frac{H^{L}}{FoV_{H}^{L}}Fov_{H}^{C}$\;

$L^{I}_{z} = \text{Zoom}(L^{I},W_{z}^{I},H_{z}^{I})$\;
Register $L^{I}_{z}$ and $L^{P}_{cy}$ using 2D MI-based method\;
Sample pixels $I^{P}_{cy}$ and $I^{I}_{z}$ from the overlapping between $L^{P}_{cy}$ and $L^{I}_{z}$\;
Recover image pixels $I^{I}_{s} = \text{DeZoom}(I^{I}_{z},W_{z}^{I},H_{z}^{I},W^{I},H^{I})$\;
Recover points $P_{s} = \text{DeSphericalProj}(I^{P}_{cy},P)$\;
$T_{init} = \text{PnPsolver}(P_{s},I^{I}_{s})$\;
\caption{Initial Calibration}
\label{alg:init}
\end{algorithm}
\subsection{Initial Calibration}\label{sec:initial}
Most targetless calibration approaches require good initialization because good initialization helps the optimization converge faster and avoid divergence or local minima. Besides, good initialization can help gradient-based optimization methods to avoid some local minimum. By utilizing the semantic information, we convert the initial calibration method into a combination of 2D image registration and Perspective-n-Point
(PnP) problem\cite{Lepetit2009}. We project LiDAR points into a spherical 2D plane \cite{Jiang2020} and get a 2D semantic range image from the LiDAR. We consider the LiDAR to be a low-resolution 360-degree camera with a partially overlapping field of view with the ordinary camera as shown in Fig.~\ref{fig:initial}. Therefore, we zoom the semantic label of the camera image into the same resolution as LiDAR using nearest neighbor interpolation method. Then, we perform 2D MI-based image registration on the two semantic images. After registration, we can get the raw correspondence between points of LiDAR and pixels of the camera image. Then, we sample several pairs of points and pixels and solve the PnP problem to get an initial calibration between the LiDAR and camera. The full algorithm is described in Algorithm.\ref{alg:init}

Because the importance of initialization, the failure of initialization will lead to the failure of the final calibration. Multiple scans will help the success of the initialization. Therefore, we propose to use multiple scans to initialize the calibration. We use the mean of the initialization parameters of the initialization results across multiple scans. In order to detect and remove the outliers of initialization parameters. We use the modified z score  proposed by Iglewicz and Hoaglin \cite{Iglewicz1993}:
\begin{equation}
    M_{i}=\frac{0.6745\left(x_{i}-\tilde{x}\right)}{\mathrm{MAD}}
\label{eq:z_score}
\end{equation}
$M_{i}$ denote the Z-score. $x_{i}$ denote the data. $\mathrm{MAD}$ denotes the median absolute deviation and $\tilde{x}$  denotes the median. We considered the initialization parameters whose absolute value of Z-scores are  greater than 3.5 as outliers\cite{Iglewicz1993}. If the outliers accounted for more than 60\%, we decide the initialization has failed.

\begin{figure}
  \centering
  \includegraphics[width=0.5\textwidth]{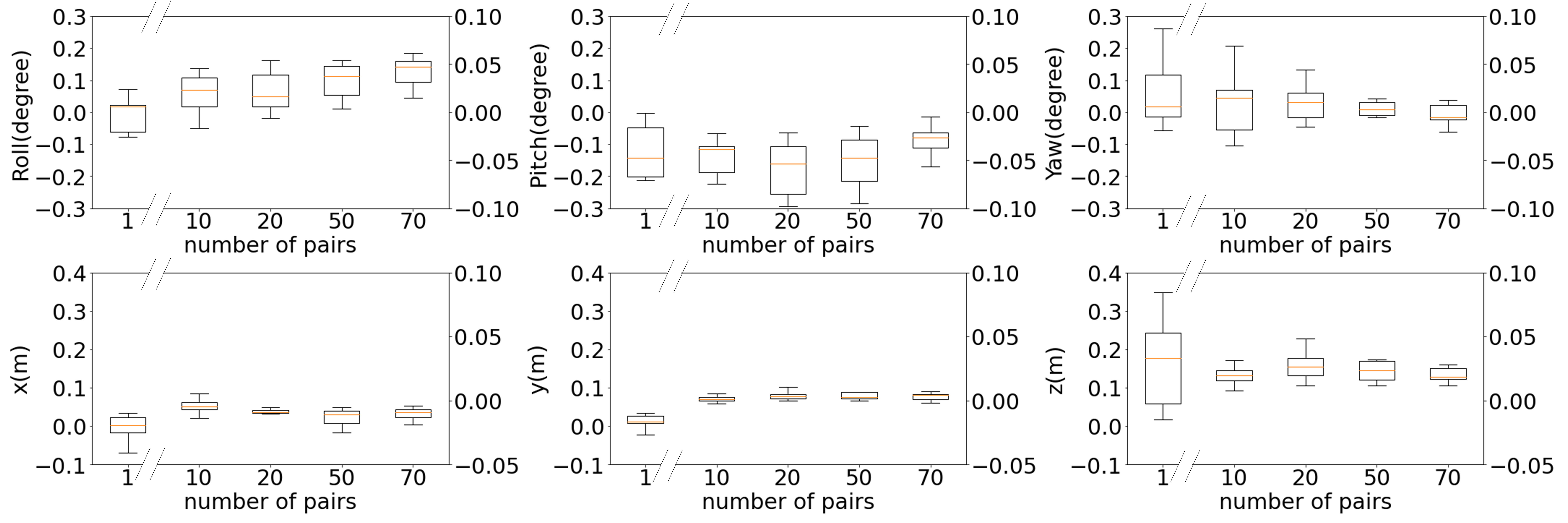}
  \caption{Calibration error for varying number of pairs of frames.}
  \label{fig:boxplot}
\end{figure}
\begin{figure}
  \centering
  \includegraphics[width=0.45\textwidth]{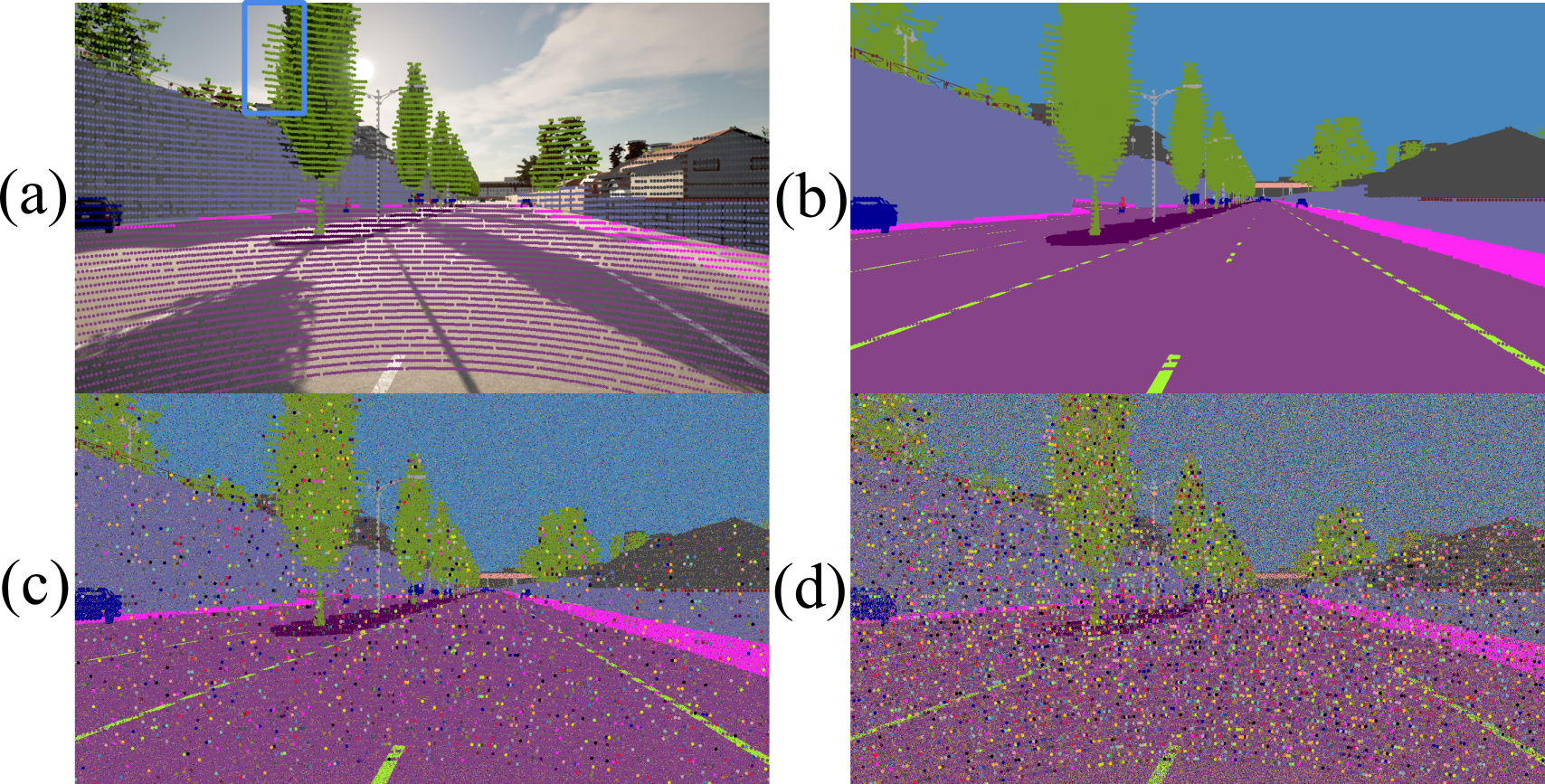}
  \caption{(a) Simulated LiDAR point cloud projected on image; (b) Ground truth image label with point cloud; (c) image label and point cloud label with 20\% error (d)  image label and point cloud label with 50\% error.}   
  \label{fig:noise_label}
\end{figure}
\section{Experiment and Results}
This section describes the experiments to evaluate the accuracy and robustness of the proposed automatic calibration technique. We first evaluate our methods on a synthetic dataset, then on the real-world KITTI360 \cite{Xie2016CVPR} and RELLIS-3D \cite{jiang2020rellis3d} datasets. 
\subsection{Synthetic dataset}
To test the accuracy and robustness of our methods, we created a dataset including paired LiDAR and camera data with semantic labels using the Carla simulator \cite{Dosovitskiy17}. 
During these experiments we used a 20 classes ontology. The image has a size of $1280\times720$. while the LiDAR has horizontal and vertical resolution of 800 and 64, respectively.

We first tested the performance and robustness of our methods and the effects of the number of data frames on overall performance. In each test, we used the ground truth labels and estimated the initial transformation through the methods described in section \ref{sec:initial}. Then, we tested the procedure with 1, 10, 20, 50, and 70 pairs of frames. As shown in Fig.~\ref{fig:boxplot}, the variance of the error decreases as we increase the number of pairs we use, which is unsurprising since by introducing more data, the estimator is able to achieve better mutual information estimation. Also, more data increase the scene's diversity and reduce the local minimum of the optimization.
\begin{figure}
  \centering
  \includegraphics[width=0.5\textwidth]{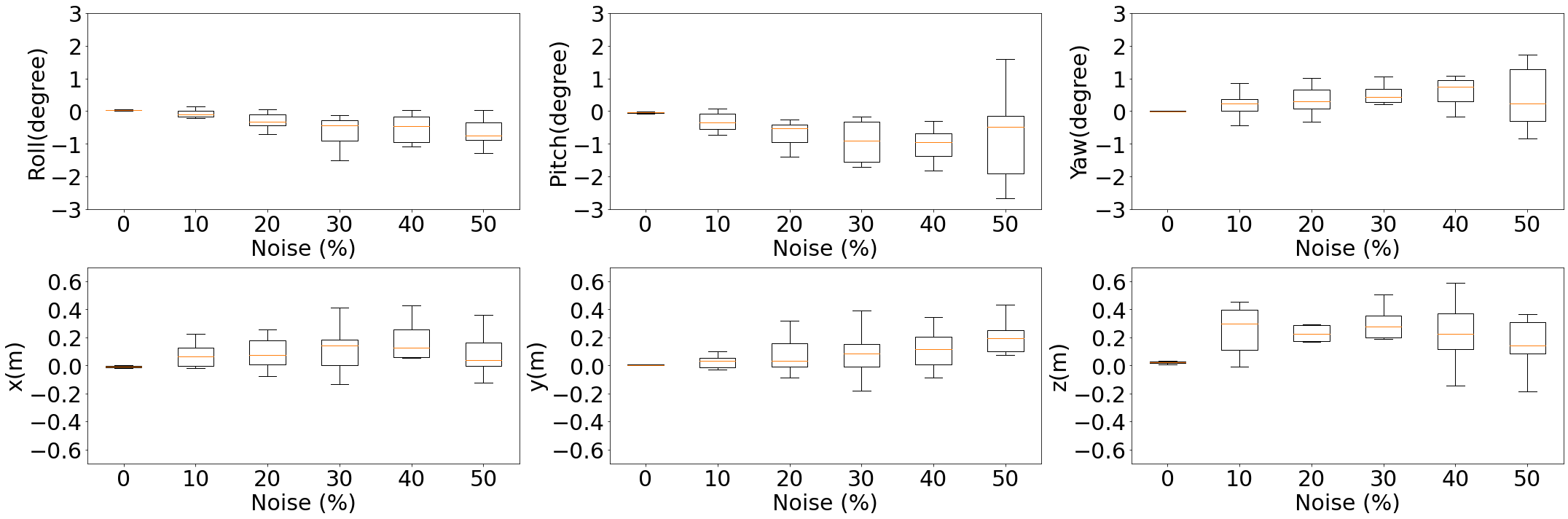}
  \caption{Calibration error for different levels of label noise}   
  \label{fig:boxplot_noise}
\end{figure}
Next, we tested the effect of noisy labels on calibration accuracy. In each test, we added random noise to both the image and point cloud labels (see Fig.~\ref{fig:noise_label}). The results are shown in Fig.~\ref{fig:boxplot_noise}. As expected, the accuracy decreases and the variance increases with increasing error noise. While we recognize the error of deep learning prediction is not Gaussian in practice, it is much easier to control the error ratio by adding noise to ground truth labels. Therefore, we use this experiment to show the effects of different levels of noise, which we supplement with experiments using model predictions in the next subsection.

\subsection{Real-world datasets}
\label{section:real_world_dataset}
While the synthetic data provides accurate ground truth transformations and semantic labels, real-world sensor and environmental characteristics are not perfectly modeled. Following our experiments with simulated data, we tested on two real-world datasets, KITTI360 and RELLIS-3D, which both provide annotated synchronized LiDAR and image data. KITTI360 dataset was collected in an urban environment, and the RELLIS-3D dataset was collected in an off-road environment. Both datasets also provide calibration parameters between LiDAR and camera. We assume them as the ground truth to test the robustness of our methods. 

We experiment with our method on human-annotated labels provided by the datasets as well as label predictions from deep learning models. For image semantic segmentation, we used the HRNet+OCR model \cite{Takikawa2019} to predict the semantic labels. For LiDAR point cloud, we used the SalsaNext model \cite{Cortinhal2020}. The RELLIS-3D dataset provides pre-trained models. The HRNet+OCR achieved 48.83\% mIoU on the RELLIS-3D dataset, and the SalsaNext SalsaNext achieved 40.20\% mIoU. The HRNet+OCR was trained using 20 classes, and the SalsaNext was trained using 16 classes due to inherent differences in detection ability of each sensor (for example, LiDAR cannot measure \texttt{sky} data). When we perform the calibration, we merge the classes into 14 common classes and ignore the \texttt{sky} class.
\begin{figure}
  \centering
  \includegraphics[width=0.5\textwidth]{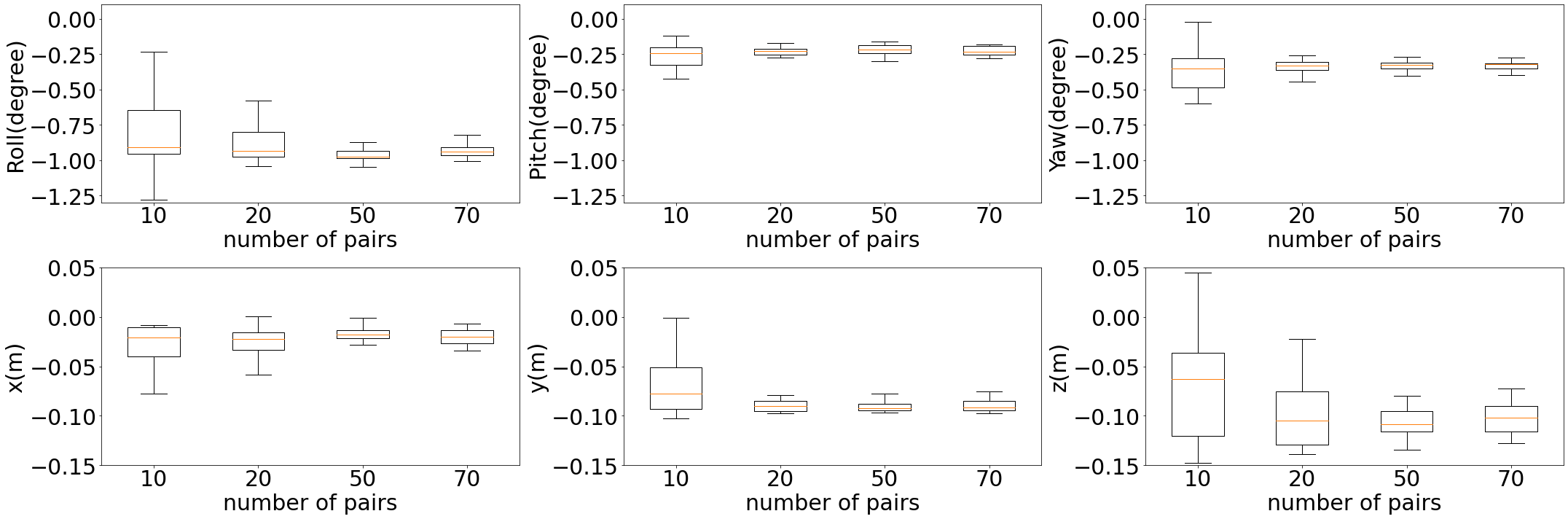}
  \caption{Calibration error using different number of pairs on RELLIS-3D Ground Truth Labels}   
  \label{fig:rellis3d_gt}
\end{figure}
\begin{figure}
  \centering
  \includegraphics[width=0.5\textwidth]{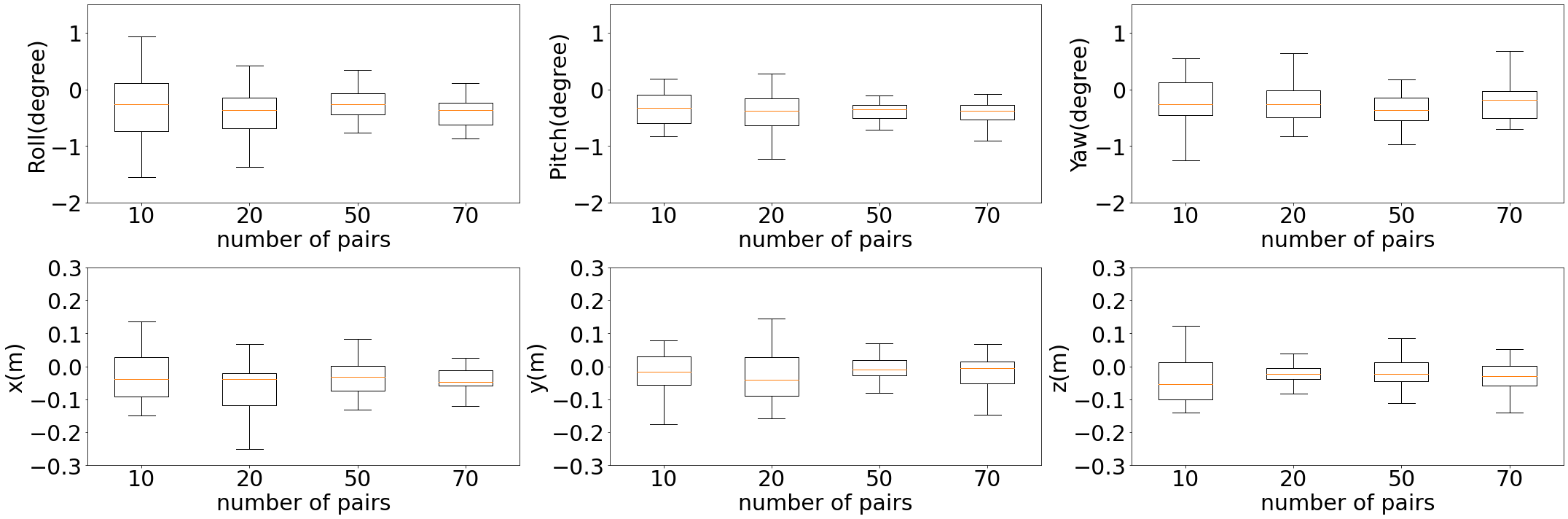}
  \caption{Calibration error using different number of pairs on RELLIS-3D Label Predictions}   
  \label{fig:rellis3d_pred}
\end{figure}
Fig.~\ref{fig:rellis3d_gt} shows the results of using ground truth labels and Fig.~\ref{fig:rellis3d_pred} shows the results of using the prediction from deep learning models. We can see that angle error using ground truth labels is within 1.5 degree using ground truth labels, and the translation error is within 0.15 meters. Unsurprisingly, the results of label prediction have larger error. Moreover, the error of the ground truth label results has an obvious bias. This might be caused by the annotation process. In RELLIS-3D, each image was annotated by humans individually, while the LiDAR scans were registered as a integrated scene and annotated together. Therefore, compared with image labels, LiDAR scan labels have better sequential consistency. Meanwhile, the labels generated by the deep learning model have better sequential consistency.   

\begin{figure}
  \centering
  \includegraphics[width=0.5\textwidth]{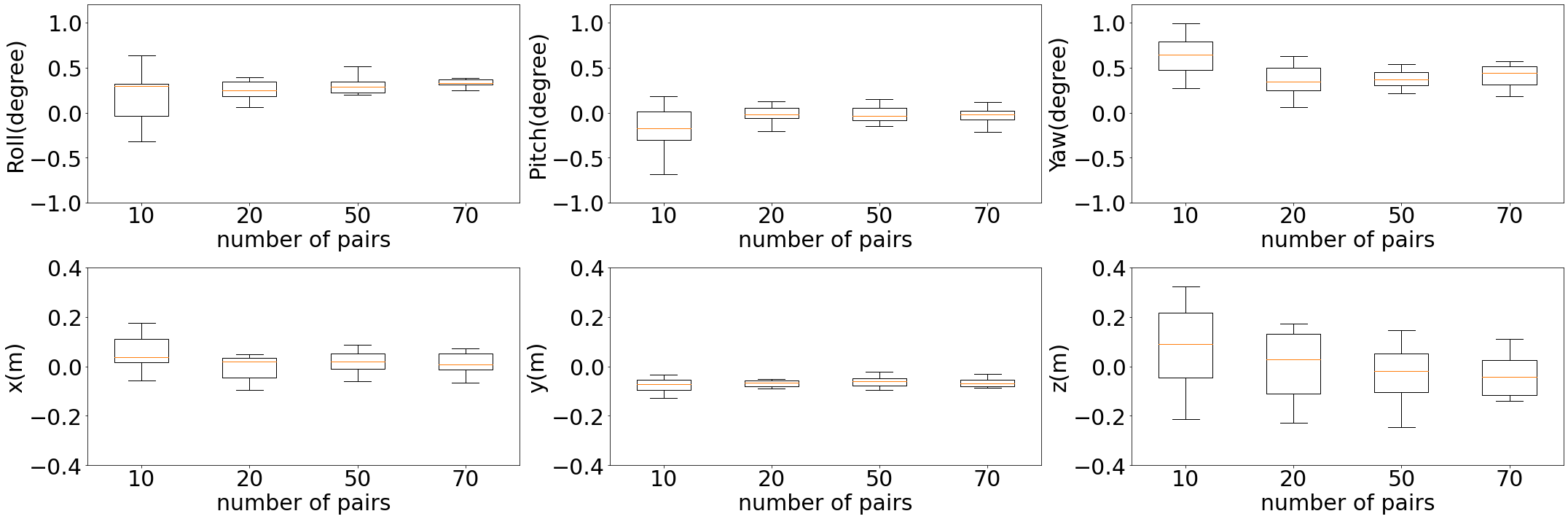}
  \caption{Calibration error using different number of pairs on KITTI360 Ground Truth Labels}   
  \label{fig:kitti360_gt}
\end{figure}

\begin{figure}
  \centering
  \includegraphics[width=0.5\textwidth]{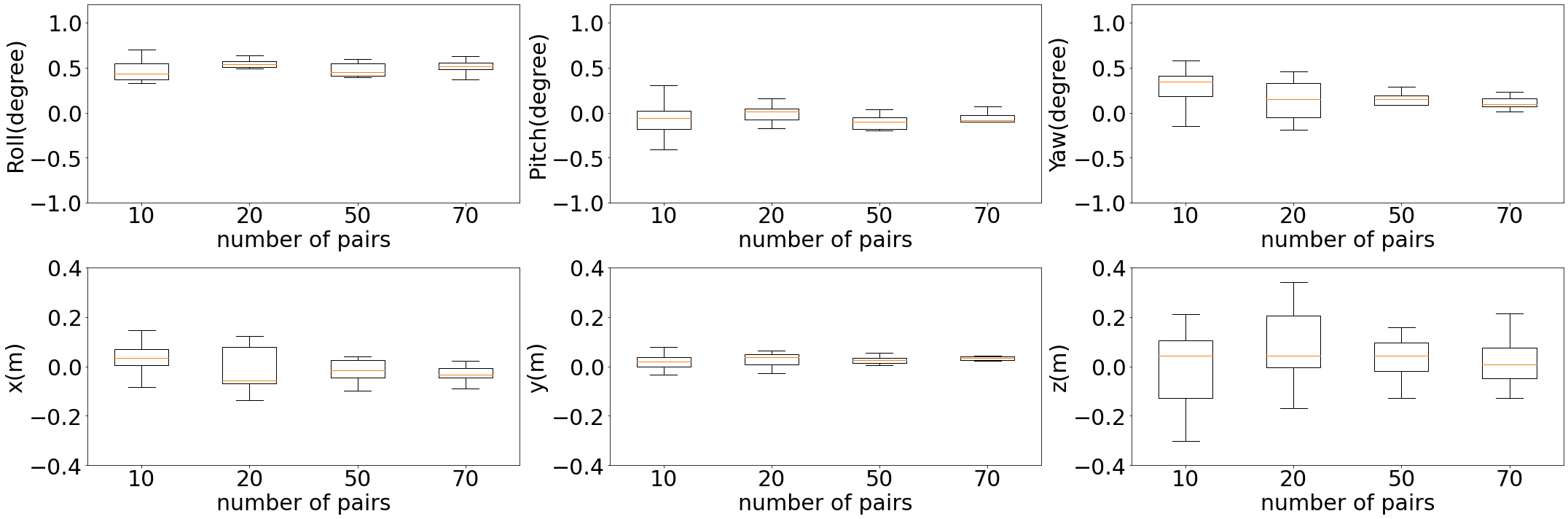}
  \caption{Calibration error using different number of pairs on KITTI360 Label Predictions}   
  \label{fig:kitti360_pred}
\end{figure}

We used the same two models to train on the KITTI360 dataset. The HRNet+OCR can achieve 31.02\% mIoU on the KITTI360 dataset and the SalsaNext can achieve 20.08\% on the KITTI360 dataset. This datasets has 45 classes which we merge into 34 for training. 
We use our proposed method to initialize the calibration and use 10, 20, 50, and 70 pairs of LiDAR-Camera data to perform calibration. Fig.~\ref{fig:kitti360_gt} shows the results of using ground truth labels and Fig.~\ref{fig:kitti360_pred} shows the results of using the prediction from deep learning models. We can see that angle error is within 1 degree predictions, and the translation error is within 0.4 meters using both of ground truth labels and label.

\begin{figure*}
  \centering
  \includegraphics[width=\textwidth]{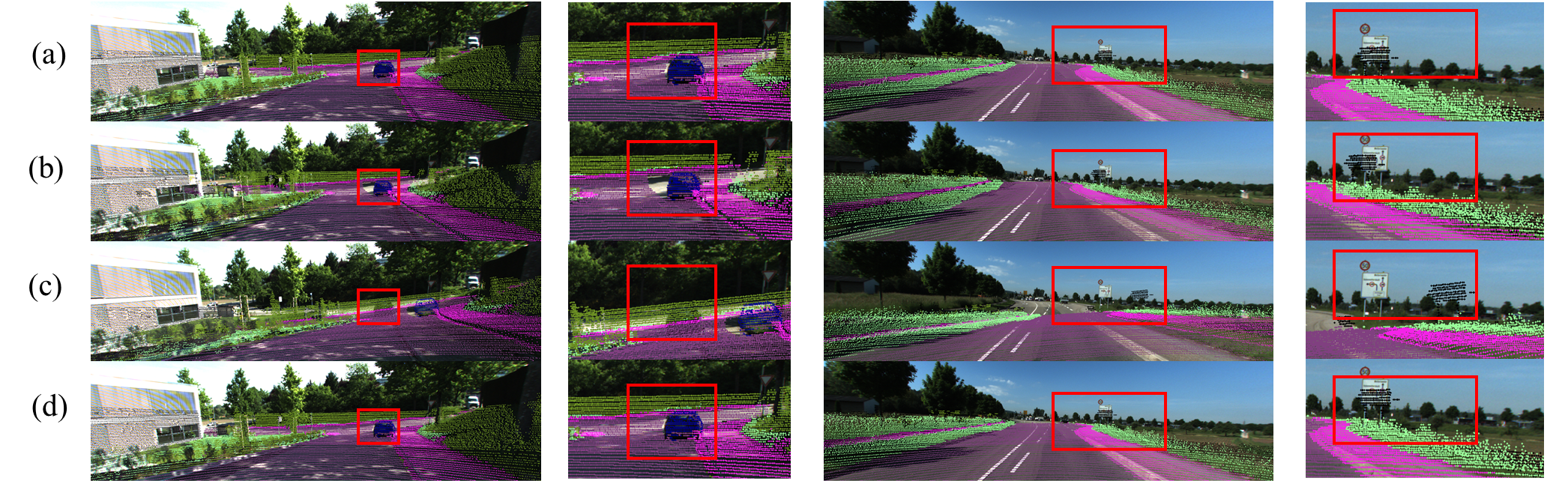}
  \caption{(a) KITT360 calibration; (b) SOIC Results; (c) PMI Results; (d) SemCal Results(Ours).  }
  \label{fig:kitti360}
\end{figure*}

\begin{figure*}
  \centering
  \includegraphics[width=\textwidth]{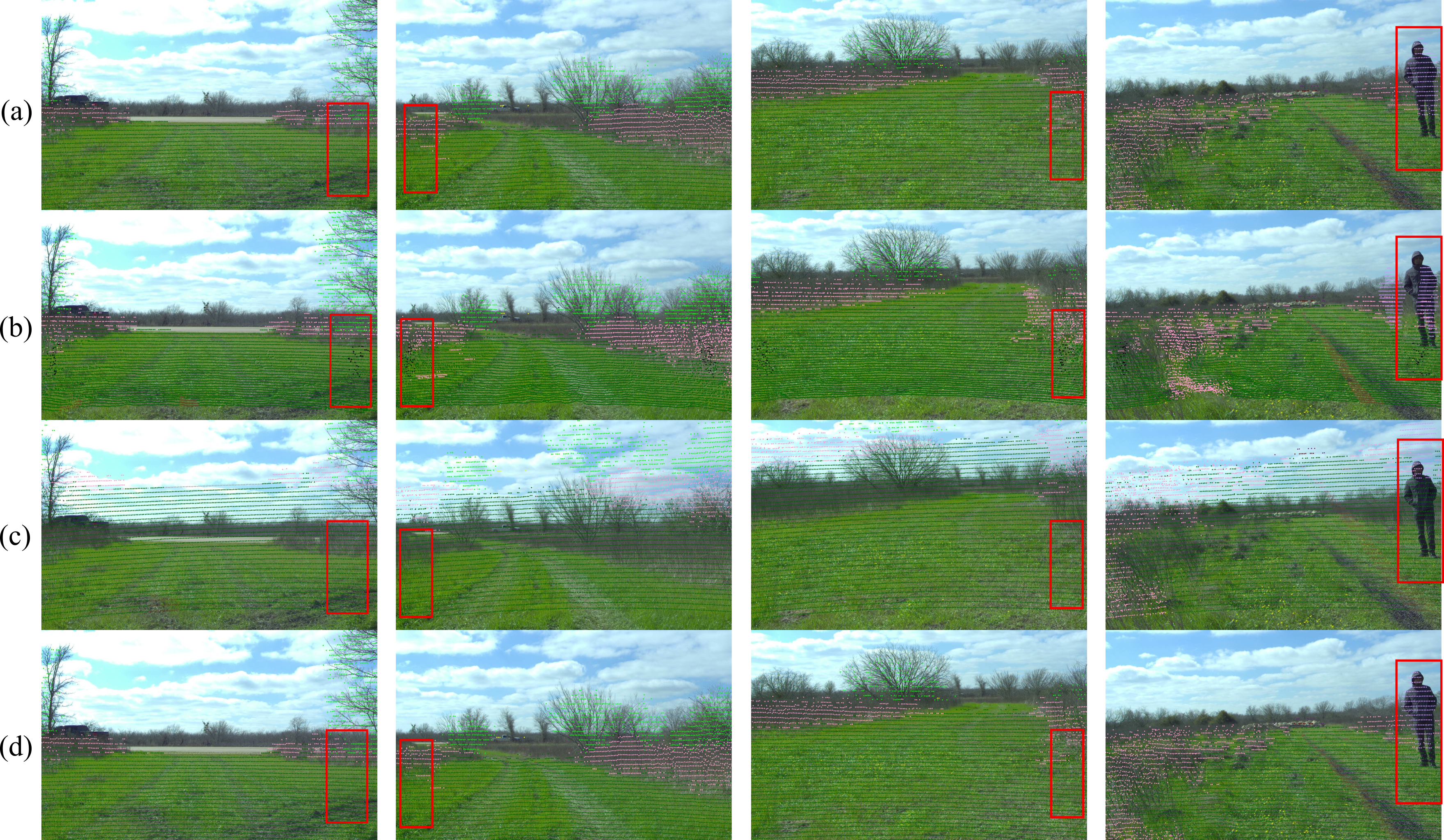}
   \caption{(a) RELLIS-3D calibration; (b) SOIC Results; (c) PMI Results; (d) SemCal Results(Ours). }
  \label{fig:rellis3d}
\end{figure*}

To evaluate against baseline calibration approaches, we compared our method with SOIC\cite{Wang2020} on the two datasets. We also implemented Pandey's \cite{Pandey2015} method to use the mutual information between the sensor-measured surface intensities to calibrate the sensors, however we use MINE\cite{Belghazi2018} to estimate the MI instead of the kernel method described in \cite{Pandey2015}. Our implementation of Pandey's method is noted as \textbf{PMI} in Table\ref{tab:kitti360} and \ref{tab:rellis3d}. All methods were initialized by our proposed method.  As shown in Table. \ref{tab:kitti360}-\ref{tab:rellis3d}, our method provide the closest results with the provided parameters of each dataset. Meanwhile, Fig.~\ref{fig:kitti360} and \ref{fig:rellis3d} shows that our calibration produces better cross-sensor data fusion than the other two methods\footnote{More visual results can be found in the video:  \url{https://youtu.be/nSNBxpCtMeo}}.  In the results of PMI, we can see there is a significant misalignment on the z-axis. This is because the grayscale image pixel value of the sky is similar to the reflectivity of the grass in LiDAR, so the method tried to align the grass ground from the LiDAR data with the sky in the camera data.
Finally, we analyzed the run time of our algorithm. Our experiment used ten labels to run the initialization algorithms and iterated the mutual information estimator, which always converged, for 4000 iterations. We ran the experiment 10 times and computed the mean run time. The initialization procedure, which only needs to be run once, took 30 seconds, and the calibration itself takes 156 seconds. Although this algorithm cannot be run in real-time, it is still fast enough to run online as a background process.


\begin{table}[]
\caption{Calibration results of KITTI360}
\centering
\begin{tabular}{  c | c | c | c | c | c | c } 
\hline
Methods &  Roll($^{\circ}$) &  Pitch($^{\circ}$) & Yaw($^{\circ}$) & X(m) & Y(m)  & Z(m) \\ 
\hline
KITTI360 &  73.72&  -71.37&64.56 & 0.26& -0.11  &  -0.83 \\ 
\hline
SOIC & 74.08 &  \textbf{-71.21} &  64.75 & 0.11 & \textbf{-0.12} &  -1.29 \\ 
\hline
PMI &  71.19 & --64.38 &  58.35 & 2.28 & -0.01 & -0.70\\ 
\hline
Ours &  \textbf{73.86} & -71.54 &  \textbf{64.69} &  0.23 &   -0.17 & \textbf{-0.89}\\ 
\hline
\end{tabular}
    \label{tab:kitti360}
\end{table}

\section{Summary and Future Work}
This paper presents a fully differential LiDAR-Camera calibration method using semantic information from both sensor measurements by optimizing mutual information between semantic labels. By utilizing semantic information, the method does not need specific targets or hand-measured initial estimates. We experiment with our method on a synthetic dataset and two real-world datasets to show our method's accuracy and robustness. 
Because the method is fully differentiable, it can be implemented using popular gradient descent optimization frameworks.
Moreover, mutual semantic information was introduced used to register multi-modal data, but the method has the potential to leverage sub-semantic intermediate features of deep networks to calibrate LiDAR and camera.
\begin{table}[]
    \centering
\caption{Calibration results of RELLIS-3D}
    \begin{tabular}{ c | c | c | c | c | c | c } 
    \hline
Methods &  Roll($^{\circ}$) &  Pitch($^{\circ}$) & Yaw($^{\circ}$) & X(m) & Y(m)  & Z(m) \\ 
    \hline
    RELLIS3D &  70.80 &  67.78 & -68.06 & -0.04 & -0.17 & -0.13\\ 
    \hline
    SOIC & 70.08 &   68.62 & -66.68 & 0.00 & -0.15 &  4.41 \\ 
    \hline
    PMI & 75.65 &  69.87 & -64.24 & -0.08 &  -0.33 & \textbf{-0.10} \\ 
    \hline
    Ours &  \textbf{70.24}&  \textbf{67.63} & \textbf{-67.32} & \textbf{-0.05} & \textbf{-0.19} & -0.06 \\ 
    \hline
    \end{tabular}
    \label{tab:rellis3d}
\end{table}





\bibliographystyle{IEEEtran}
\bibliography{IEEEabrv,references}

\end{document}